\documentclass[11pt, conference, letterpaper]{IEEEtran}
\IEEEoverridecommandlockouts
\usepackage{amsmath,amssymb,amsfonts}
\usepackage{algorithmic}
\usepackage{algorithm}
\usepackage{textcomp}
\usepackage{xcolor}
\usepackage[T1]{fontenc}      
\usepackage{booktabs}       
\usepackage{nicefrac}       
\usepackage{microtype}      
\usepackage{xcolor}         
\usepackage{graphicx}
\usepackage{subcaption}
\usepackage{placeins}
\usepackage[a4paper,
        bindingoffset=0.24in,
        left=.6in,
        right=.6in,
        top=1.95cm,
        bottom=4.45cm,
        footskip=.25in]{geometry}  
\begin{document}

\title{SPEAR: Soft Prompt Enhanced Anomaly Recognition for Time Series Data\\
}

\author{
\IEEEauthorblockN{Hanzhe Wei, Jiajun Wu, Jialin Yang, Henry Leung, and Steve Drew}
\IEEEauthorblockA{
\textit{Department of Electrical and Software Engineering}\\
\textit{University of Calgary, Calgary, Canada}\\
\{hanzhe.wei, jiajun.wu1, jialin.yang, leungh, steve.drew\}@ucalgary.ca}
}


\maketitle

\begin{abstract}
Time series anomaly detection plays a crucial role in a wide range of fields, such as healthcare and internet traffic monitoring. The emergence of large language models (LLMs) offers new opportunities for detecting anomalies in the ubiquitous time series data. Traditional approaches struggle with variable-length time series sequences and context-based anomalies. We propose Soft Prompt Enhanced Anomaly Recognition (SPEAR), a novel approach to leverage LLMs for anomaly detection with soft prompts and quantization. Our methodology involves quantizing and transforming the time series data into input embeddings and combining them with learnable soft prompt embeddings. These combined embeddings are then fed into a frozen LLM. The soft prompts are updated iteratively based on a cross-entropy loss, allowing the model to adapt to time series anomaly detection. The use of soft prompts helps adapt LLMs effectively to time series tasks, while quantization ensures optimal handling of sequences, as LLMs are designed to handle discrete sequences. Our experimental results demonstrate that soft prompts effectively increase LLMs' performance in downstream tasks regarding time series anomaly detection. 
\end{abstract}


\section{Introduction}
The promise of heading to artificial general intelligence (AGI)  with large language models (LLMs) has inspired researchers to explore its use for time series analysis in domains such as healthcare, industrial systems monitoring, and satellite telemetry monitoring \cite{zhou2023one, jin2023time, li2022evaluating, xue2022leveraging, ott2021robust, shao2022log}. The auto-regressive features of LLMs allow them to be effective for time-series forecasting tasks \cite{gruver2024large} and even the more challenging anomaly detection tasks. 
Most existing approaches for using LLMs as anomaly detectors for time series data fall under three categories: prompt-based \cite{xue2023promptcast, liu2024logprompt}, fine-tuning \cite{dang2021ts, xue2022leveraging, dang2020time}, and zero-shot \cite{gruver2024large, alnegheimish2024large}. However, these methods require carefully crafted chat templates for LLMs to produce the desired response \cite{alnegheimish2024large}. Fine-tuning LLMs for improved performance presents significant challenges, primarily due to the substantial computational costs involved \cite{su2024large} and possible distortion of pretrained features \cite{kumar2022fine}. Using LLMs for zero-shot anomaly detection typically requires models trained on vast amounts of data or advanced models like GPT-4, which brings substantial financial costs and privacy implications \cite{alnegheimish2024large}.
However, computational cost and response time are crucial in real-world applications, such as healthcare and Internet-of-things (IoTs). Therefore, larger frontier LLMs might not always be the best option.

Recently, soft prompts have gained popularity as an efficient way to adapt pretrained LLMs to specific tasks without crafting detailed prompts \cite{lu2023decomposed, vu2021spot, wu2022adversarial, qin2021learning, huang2023soft}. Applying LLMs for domain and dataset-specific tasks can be highly time-consuming and resource-intensive. As a result, effectively utilizing smaller language models to achieve or even surpass the performance of large-scale LLMs without the overhead of training or fine-tuning represents a critical research bottleneck. Smaller LLMs paired with soft prompts offer several advantages in terms of computational efficiency, reduced cost, and improved privacy since they can be easily deployed locally. Leveraging these Smaller LLMs can substantially lower barriers to entry, enabling broader adaptation across massive real-world applications. This approach aligns well with the growing demand for scalable, privacy-preserving solutions.

In summary, our work provides the following contributions:
\begin{itemize}
    \item In this paper, we pioneer the work of pairing smaller LLMs with soft prompts in an attempt to achieve performance levels comparable to larger SOTA models. 
    \item We propose \textbf{Soft Prompt Enhanced Anomaly Recognition} (SPEAR), which adapts small LLMs for time series anomaly detection, demonstrating the feasibility of leveraging pre-trained LLMs in data analysis tasks beyond simple NLP tasks. To our knowledge, SPEAR is the first framework leveraging soft prompts to adapt small-sized LLMs to detect time series anomalies.
    \item We show that soft prompts can significantly enhance LLM performance on time series tasks without fine-tuning the entire model. We demonstrate that using soft prompts eliminates the need to construct chat templates or employ various prompting techniques. We evaluate the performance of SPEAR  across three datasets: MIMIC-IV \cite{johnson2023mimiciv}, NAB \cite{lavin2015evaluating}, and NASA datasets \cite{hundman2018detecting}, and SPEAR outperforms LLMs without soft prompts in a zero-shot scenario and requires no additional prompting.
\end{itemize}

\section{Related Works}
\label{relatedworks}

\begin{figure*}[!htb]
  \centering
  \includegraphics[width=1\linewidth]{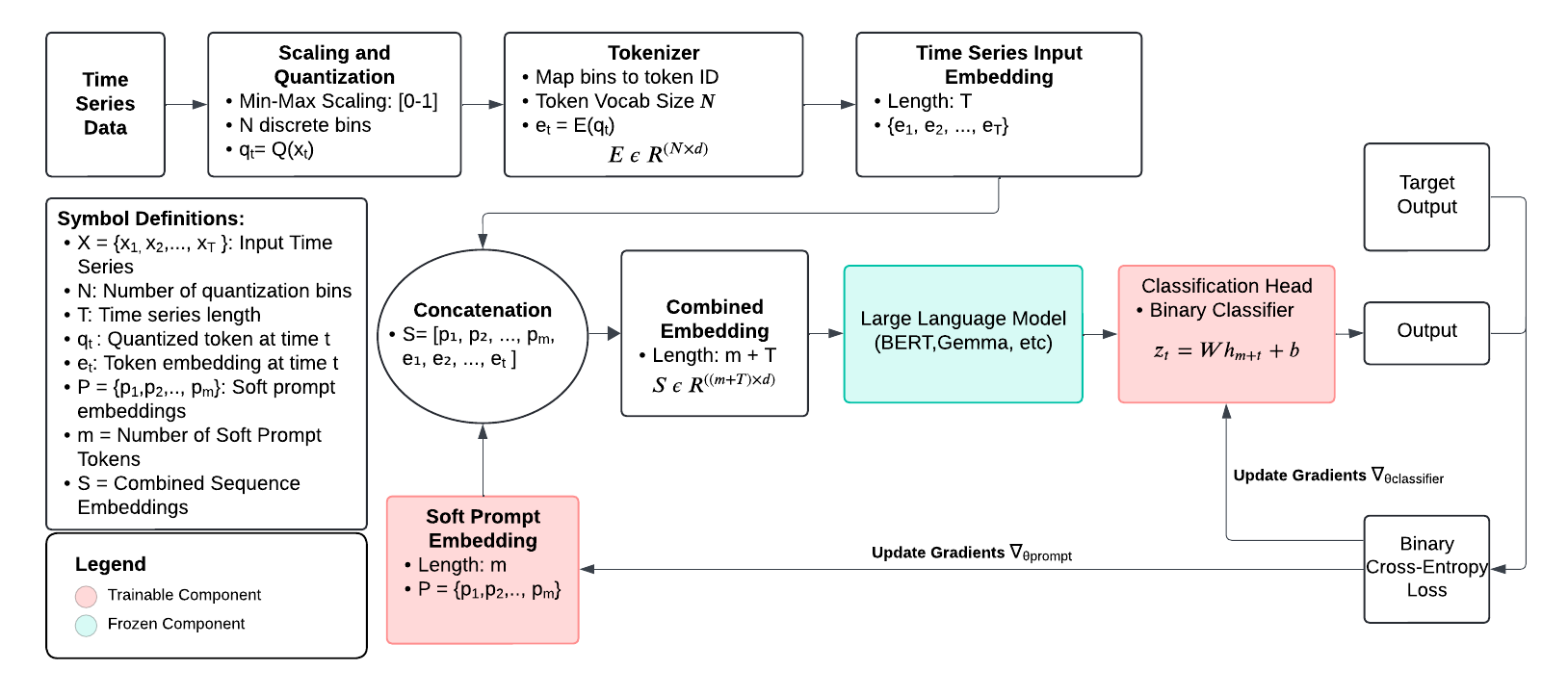} 
  \caption{Soft Prompt Enhanced Anomaly Recognition for Time Series Architecture Diagram.}
  \label{fig:spear-ts-architecture}
\end{figure*}

\subsection{Soft Prompts} Soft prompts have emerged as an efficient method for adapting LLMs to specific tasks without fine-tuning the entire model. Lester et al. \cite{lester2021power} introduced the concept of \textit{prompt tuning}, showing that by only tuning continuous prompt embeddings, performance comparable to full fine-tuning can be achieved. This approach has been successfully applied in various natural language processing tasks but has not been extensively explored in the context of time series data \cite{lu2023decomposed, wu2022adversarial, vu2021spot}.

\subsection{LLMs for Anomaly Detection}
Recent work has begun to explore using LLMs for anomaly detection tasks. Zhang et al. \cite{zhang2023llmad} proposed LLMAD, a method that leverages LLMs for time series anomaly detection using few-shot learning and a chain-of-thought approach. However, their method differs from our SPEAR because it does not utilize soft prompts or quantization techniques. Similarly, Liu et al. proposed a method employing LLMs for accurate and interpretable time series anomaly detection, achieving performance comparable to state-of-the-art deep learning methods while offering enhanced interpretability \cite{liu2024logprompt}. Alnegheimish et al. \cite{alnegheimish2024large} were able to convert signals into text, enabling LLMs to work with time series data and perform better than traditional methods such as ARIMA and LSTM.

While these results are promising, especially for interpretability and flexibility, the complexity of LLMs becomes particularly valuable in environments where traditional models struggle to scale or adapt to highly heterogeneous, context-rich data. In complex and dynamic domains, such as industrial manufacturing systems, cybersecurity, and computational workflows, LLMs' advanced semantic reasoning and contextual awareness offer significant advantages over simpler models. For example, Russel et al. introduced the AAD-LLM framework, which leverages LLMs’ ability to integrate domain knowledge and adapt to evolving patterns, making it effective for detecting context-dependent anomalies without retraining \cite{russell2024aad}. In the cybersecurity domain, TAD-GP utilizes fine-tuned LLMs to detect complex network anomalies, outperforming conventional methods on benchmark datasets while maintaining computational efficiency \cite{zhao2025efficient}. Similarly, in computational science, LLMs have been shown to detect workflow execution anomalies by generalizing from a small number of examples through in-context learning and fine-tuning \cite{jin2024large}.

These use cases emphasize that LLMs are justified and important to consider in settings where data is noisy, highly variable, or contains rich context not easily captured by statistical or traditional deep learning methods. Their ability to generalize from limited data, interpret sequential information, and incorporate domain-specific knowledge via prompting makes LLMs valuable in modern anomaly detection pipelines, especially in places where data often exhibits irregular sampling and complex temporal dependencies.

\section{Methodology}
\label{method}

SPEAR is designed to leverage LLMs' feature extraction and contextual understanding capabilities for time series anomaly detection. It consists of three components: a pre-trained LLM, soft prompts for task-specific fine-tuning, and quantization for efficient computation. The overall process can be summarized as follows: 1. Time series data is preprocessed, quantized, and encoded into a format suitable for LLM input. 2. Soft prompts are initialized and optimized for the anomaly detection task. 3. The LLM with soft prompts processes the input data. 4. The output is used to identify anomalies.

\subsection{Framework} The SPEAR approach begins with the preprocessing of raw time series data with scaling and quantization. Quantization converts the continuous time series data \(X = \{x_1,x_2,...,x_T\}\) into a sequence of discrete tokens. The range of the time series data is divided into \(N\) discrete bins(note that \(N\) is different for every dataset). The data is transformed into discrete data \(q_t = Q(x_t)\), making it more compatible with token-based models like LLMs. 
\begin{equation}
 \label{eq1}
 Quantization\ Bins = \{q_1, q_2,.., q_N\} 
\end{equation}

This preprocessed data is then tokenized into embeddings where each quantized token \(q_t\) is mapped to a high-dimensional embedding vector \(e_t\) to capture semantic relationships. An embedding matrix \(E \in \mathbb{R}^{N\times d} \), where \(d\) is the embeddings dimension. The input tokens are mapped into embeddings like \(e_t = E_{q_t}\) for \(t = 1,2,..., T\).  
Then, soft prompts are introduced as learnable embedding vectors that guide the LLM's attention towards the anomaly detection task without modifying the LLM's weights. First, the sequence of learnable embeddings is initialized as \(P = {p_1,p_2,...,p_m}\). During training, the soft prompts \(P\) are updated via backpropagation. 

The soft prompt embeddings and the input embeddings are combined to form a single input sequence for the LLM. \(S=[p_1,p_2,...,p_m, e_1, e_2, ..., e_T]\). This step of combining the embeddings is important as it allows the soft prompts to influence the processing of the input embeddings throughout the LLM's layers.

The combined embedding is then fed into a pre-trained, frozen LLM (such as BERT, Gemma, etc). The LLM processes this input and produces the output embeddings. A small classification head is added to the LLM to perform binary classification. A linear layer is added which maps the LLM's outputs to logits \(z_t\), and then a sigmoid function is used to convert the logits into probabilities \(\hat{y}_t\).

\begin{equation}
 \label{eq2}
 z_t = Wh_{m+t} + b
\end{equation}

\begin{equation}
 \label{eq3}
 \hat{y}_t = \sigma(z_t) = \frac{1}{1+e^{-z_t}}
\end{equation}

Notably, while the LLM remains frozen, the soft prompt embeddings are being updated based on the binary cross-entropy loss calculated between the model's output and the target label shown in Fig.~\ref{fig:spear-ts-architecture}. The loss function computes the difference between the predicted and true labels; the gradients of this cross-entropy loss are used to update only the soft prompt embeddings through the Adam optimizer. SPEAR allows the fine-tuning of soft prompts on LLMs on time series anomaly detection without altering the underlying LLM.

\subsection{Task} We approach anomaly detection as a binary classification task within time series data. Our goal is to categorize individual data points or sequences as either normal or anomalous across diverse datasets. By employing our SPEAR method on these varied time series, we seek to showcase its robustness and efficacy in identifying anomalies across multiple domains and data types. This binary classification framework allows us to pinpoint unusual patterns or outliers that may signify critical events or deviations.

\subsection{Data preprocessing.} We address two critical challenges in our preprocessing stage: class imbalance and variable sequence lengths. To mitigate the issue of imbalanced data, we employ the Time Series-Synthetic Minority Over-sampling Technique (T-SMOTE) \cite{pears2014synthetic}. T-SMOTE generates synthetic examples of the minority class, effectively balancing our dataset and preventing model bias towards the majority class. This is particularly crucial in anomaly detection, where anomalous instances are commonly underrepresented. Additionally, our dataset contains sequences ranging from 1 to 100 time steps in length. Longer sequences are truncated to fit within the model's capacity. This approach enables us to maintain a consistent input format across all samples, allowing for efficient batch processing and model training while preserving as much original information as possible.

\noindent\textbf{Context Anomalies Labeling.}
\label{app:context-anomaly} MIMIC-IV data initially only contained point anomalies; additional context-based anomalies were introduced by us to enrich the dataset. These context-based anomalies consider the temporal relationships and patterns within the data, often providing more actionable insights than isolated outliers \cite{zamanzadeh2024deep}.

Given a univariate time series $\{x_1, x_2, \ldots, x_T\}$, the following statistical tests and heuristics were applied to identify different classes of contextual anomalies:

\subsubsection{Monotonic Trend Detection}

Monotonic trends represent a consistent increase or decrease in the signal over time. These are detected using simple linear regression.

Let $t = \{1, 2, \ldots, T\}$ be the time index. Fit a linear model: $x_t = \beta_0 + \beta_1 t + \epsilon$
A time series is labelled as a \textit{Monotonic Trend} if:
$|\beta_1| > 0.01 \quad \text{and} \quad p\text{-value} < 0.05$, where $\beta_1$ is the slope and the $p$-value is from a two-sided hypothesis test for $\beta_1 = 0$.

\begin{algorithm}[h]
\caption{Monotonic Trend Detection}
\label{alg:monotonic}
\begin{algorithmic}[1]
\STATE \textbf{Input:} Time series $X = \{x_1, x_2, \ldots, x_T\}$ 
\STATE \textbf{Output:} Label as `Monotonic Trend' if detected
\STATE Fit linear regression: $x_t = \beta_0 + \beta_1 t + \epsilon$
\IF{$|\beta_1| > 0.01$ \textbf{and} p-value $< 0.05$}
    \RETURN `Monotonic Trend'
\ELSE
    \RETURN None
\ENDIF
\end{algorithmic}
\end{algorithm}

\begin{algorithm}[t]
\caption{Sudden Spike Detection}
\label{alg:spike}
\begin{algorithmic}[1]
\STATE \textbf{Input:} Time series $X = \{x_1, x_2, \ldots, x_T\}$ 
\STATE \textbf{Output:} Label as `Sudden Spike' if detected
\STATE Compute first differences: $d_t = |x_t - x_{t-1}|$ for $t = 2, \ldots, T$
\STATE Compute threshold: $\theta = \mu_d + 3\sigma_d$
\IF{$\exists d_t > \theta$}
    \RETURN `Sudden Spike'
\ELSE
    \RETURN None
\ENDIF
\end{algorithmic}
\end{algorithm}

\subsubsection{Sudden Spike Detection}

Sudden spikes are brief but significant deviations in the measurement trajectory. They are detected by computing first-order differences: $d_t = |x_t - x_{t-1}|, \quad \text{for } t = 2, \ldots, T$

A spike is flagged if: $\exists t \in \{2, \ldots, T\}$ $\text{ such that } d_t > \mu_d + 3\sigma_d,$ where $\mu_d$ and $\sigma_d$ are the mean and standard deviation of $\{d_t\}$.

\subsubsection{Sudden Shift Detection}

A sudden shift represents a change in the baseline level of the signal. This is detected by iteratively splitting the time series at different points and conducting a two-sample t-test: \text{Split } $\{x_1, \ldots, x_T\}$ \text{ into } 
$A = \{x_1, \ldots, x_k\}, B = \{x_{k+1}, \ldots, x_T\}$. A shift is detected if for any valid $k$:\text{p-value from } t\text{-test}(A, B) < 0.05.

\subsubsection{Volatility Change Detection}

This anomaly captures a shift in variance (volatility) between two segments of the time series. Detection is based on Levene’s test for equal variances: $\text{Split as before: } A = \{x_1, \ldots, x_k\}, B = \{x_{k+1}, \ldots, x_T\}.$ Flag a \textit{Volatility Change} if: p-value from Levene’s test$(A, B) < 0.05.$

\begin{algorithm}[H]
\caption{Volatility Change Detection}
\label{alg:volatility_ieee}

\begin{algorithmic}[1]
\STATE \textbf{Input:} Time series $X = \{x_1, x_2, \ldots, x_T\}$ 
\STATE \textbf{Output:} Label as `Volatility Change' if detected
\FOR{each split point $k \in [2, T-2]$}
    \STATE Split series: $A = \{x_1, \ldots, x_k\},$ 
    \STATE $B = \{x_{k+1}, \ldots, x_T\}$
    \STATE Perform Levene's test on $A$ and $B$
    \IF{p-value $< 0.05$}
        \RETURN `Volatility Change'
    \ENDIF
\ENDFOR
\RETURN None
\end{algorithmic}
\end{algorithm}

While these statistical methods for injecting context anomalies provide a simple and interpretable approach, we recognize their limitations in fully capturing the complexity of real-world clinical anomalies. These patterns should be viewed as proxy anomalies rather than medically validated events. Future work will involve collaboration with clinical experts to validate and refine anomaly labels using insights from the domain.

\textbf{T-SMOTE}
\label{app:T-SMOTE}
The Time Series Synthetic Minority Over-sampling Technique (T-SMOTE) is an adaptation of the well-known Synthetic Minority Over-sampling Technique (SMOTE) algorithm, designed to address the challenges of imbalanced datasets in time series data. T-SMOTE addresses the issue of class imbalance by generating synthetic samples for the minority class in time series data, thereby helping to balance the dataset and enhance model training. The following steps detail how T-SMOTE works in practice:

\begin{enumerate}
    \item \textbf{Neighbours Selection}: T-SMOTE identifies k-nearest neighbours (usually done with Euclidean distance or other distance metrics) of each sample in the minority class. The neighbours are selected from the minority class, ensuring the synthetic samples remain representative of this class.
    \item \textbf{Interpolation}: For each minority sample $x_i$, T-SMOTE randomly selects one of its nearest neighbors $x_{nn}$, and generates a new synthetic time series point $x_{syn}$ by interpolating between $x_i$ and $x_{nn}$, using the formula: 
        \begin{equation}
        x_{syn} = x_i + \lambda \times (x_{nn} - x_i)
        \end{equation}
    where $\lambda$ is a random value between 0 and 1. This generates a point that lies along the line connecting $x_i$ and $x_{nn}$, effectively creating new time series samples.
\end{enumerate}

Here is an example of how T-SMOTE works: 
Let's assume there is a time series dataset with two classes, where each sample contains six data points. The majority class is the normal class, and the minority class is the anomalous class. The dataset is highly imbalanced, taking two minority class samples: 
Minority class sample 1: $x_1 = [4,3,4,8,7,3]$, and Minority class sample 2: $x_2 = [3,4,4,3,7,4]$


After applying T-SMOTE, $x_{syn}$ is obtained with its values being $[3.5,3.5,4,5.5,7,3.5]$. 

While T-SMOTE is a useful method for addressing class imbalances in time series data, it has limitations. Since T-SMOTE generates synthetic samples by interpolating between existing minority class examples, it may inadvertently create samples that are too similar to the original ones, which may lead to overfitting. Although T-SMOTE attempts to preserve temporal structure, simple interpolation may not fully capture the underlying time-dependent dynamics, especially in complex time series data where important features (e.g., trends, seasonality) may be lost. 

This study utilizes the T-SMOTE methodology for NASA and NAB AWS datasets, both are highly imbalanced classes.  

\section{Experimental Results}
\label{experiments}

\vspace*{0.1in}
\begin{table*}
  \caption{Evaluation Results for Anomaly Detection of Time Series in MIMIC IV Dataset}
  \label{tab:MIMIC_IV_eval}
  \centering
  \begin{tabular}{ccccccc}
    \toprule
    Method                      & Accuracy  & F1-Score & Recall & Precision & AUROC & AUPR \\
    \midrule
    Gemma Zero Shot             & 0.5450 & 0.7774 & 0.5800 & 0.5421 & 0.5457 & 0.5611 \\
    GPT4 Zero Shot              & 0.6500 & 0.6789 & 0.7400 & 0.6271 & 0.6956 & 0.6836 \\
    BERT Zero Shot    & 0.4900 & 0.1356 & 0.0800 & 0.4444  & 0.0760 & 0.2622\\
    LSTM      & 0.7700 & 0.7767 & 0.8000 & 0.7547 &0.7960 & 0.7774\\ 
    \midrule
    SPEAR-BERT(\textbf{Ours}) & \textbf{0.9300} & \textbf{0.9285} & \textbf{0.9100} & \textbf{0.9479} &\textbf{0.9323} &\textbf{0.9290} \\
    SPEAR-Gemma(\textbf{Ours}) & 0.6800  & 0.7090 &  0.7800 & 0.6500 & 0.7254 & 0.7150 \\
    \bottomrule
  \end{tabular}
\end{table*}

\begin{table*}
  \caption{Evaluation Results for Anomaly Detection of Time Series in NASA Satellite Dataset}
  \label{tab:NASA_eval}
  \centering
  \begin{tabular}{ccccccc}
    \toprule
    Method                      & Accuracy  & F1-Score & Recall & Precision & AUROC & AUPR \\
    \midrule
    Gemma Zero Shot             & 0.1834 & 0.1266  & 1.0000 & 0.0676  & 0.5660 & 0.0676 \\
    GPT4 Zero Shot              & 0.2367 & 0.1342 & 1.0000 & 0.0719 & 0.5943 & 0.0719 \\
    BERT Zero Shot              & 0.0592 & 0.1117 & 1.0000 & 0.0592 & 0.5000 & 0.0592 \\
    LSTM      & 0.8714 & 0.4000 & 0.4286 & 0.3750 & 0.8549 & 0.3445\\ 
    \midrule
    SPEAR-BERT(\textbf{Ours}) & \textbf{0.9112} & \textbf{0.4000} & \textbf{0.5000} & \textbf{0.3333} &\textbf{0.6157} &\textbf{0.5039} \\
    SPEAR-Gemma(\textbf{Ours}) & 0.7515 & 0.2222 &  0.6000 & 0.1364 & 0.7371 & 0.2806 \\
    \bottomrule
  \end{tabular}
\end{table*}

\begin{table*}
  \caption{Evaluation Results for Anomaly Detection of Time Series in NAB AWS Dataset}
  \label{tab:NAB_eval}
  \centering
  \begin{tabular}{ccccccc}
    \toprule
    Method                      & Accuracy  & F1-Score & Recall & Precision & AUROC & AUPR \\
    \midrule
    Gemma Zero Shot             & 0.4583 & 0.0714  & 0.7778 & 0.0374 & 0.6694 & 0.0486 \\
    GPT4 Zero Shot              & 0.5357 & 0.0714 & 0.6667 & 0.0377 & 0.5994 & 0.0341 \\
    BERT Zero Shot              & 0.1265 & 0.0578 & 1.0000 & 0.0298 & 0.6616 & 0.0546 \\
    LSTM      & 0.8935 & 0.0317 & 0.0588 & 0.0217 & 0.5076 & 0.0872 \\ 
    \midrule
    SPEAR-BERT(\textbf{Ours}) & \textbf{0.8512} & \textbf{0.1071} & \textbf{0.3333} & \textbf{0.0638} &\textbf{0.7101} &\textbf{0.0568} \\
    SPEAR-Gemma(\textbf{Ours}) & 0.6101 & 0.0775 & 0.6111  & 0.0414 & 0.6823 & 0.0499 \\
    \bottomrule
  \end{tabular}
\end{table*}

\begin{figure}[!h]
  \centering
  \includegraphics[width=1\linewidth]{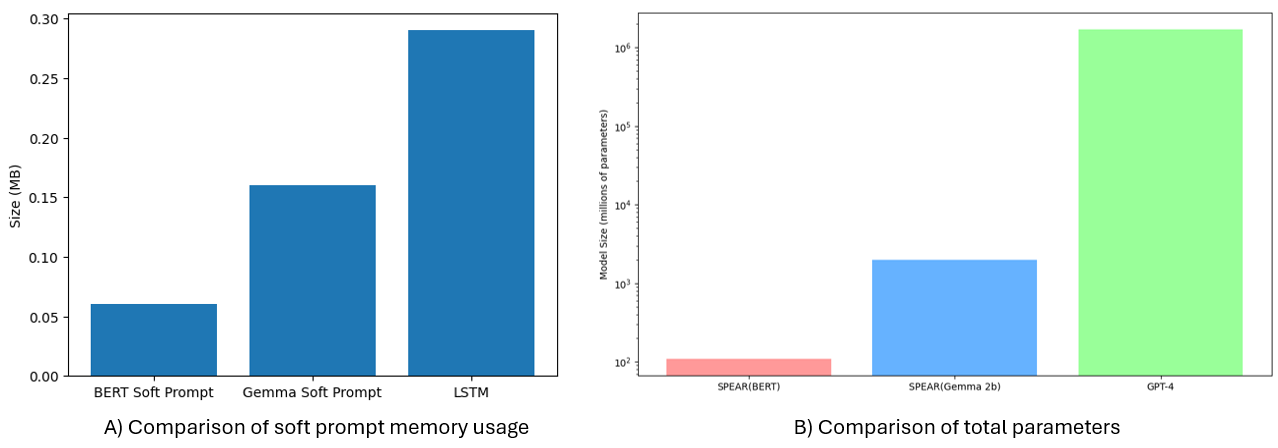} 
  \caption{Memory Usage and Parameters Comparisons}
  \label{fig:soft_prompt_sizes}
\end{figure}
\subsection{Datasets} To evaluate the efficacy of SPEAR, we conduct experiments on three different datasets across various domains. Our experiments utilized MIMIC-IV \cite{johnson2023mimiciv}, NASA's satellite telemetry \cite{hundman2018detecting}, and the Numenta Anomaly Benchmark (NAB) \cite{lavin2015evaluating}.
MIMIC-IV comprises the numerical results of patients' historical medical labs. NASA Satellite Telemetry consists of measurements from spacecraft systems, presenting unique challenges in detecting anomalies in critical space operations. Numenta Anomaly Benchmark (NAB) is a diverse collection of real-world and artificial time series data designed to test and compare anomaly detection algorithms across various scenarios. 

For MIMIC-IV, we focused on the ten most common lab exams, with input comprising a patient's sequential lab tests across their medical history. However, the data in MIMIC-IV initially only contains point anomalies, where individual data points falling outside of the acceptable range are flagged as anomalies. Additional context-based anomalies were introduced and labeled to enrich the time series data and make it more representative of real world scenarios (the full algorithm to label context anomalies is explained in ~\ref{app:context-anomaly}). 
The NASA dataset was processed by splitting sequences into windows of 100 data points with a stride of 10, evaluating each telemetry stream individually. For NAB, we used variable window sizes ranging from 1 to 50 data points, evaluating each dataset separately. These preprocessing steps were designed to capture the unique temporal characteristics of each dataset while ensuring the flexibility of input formats for our SPEAR model across different types of time series data.




\subsection{Baseline Comparisons} We compare SPEAR with several baselines: LSTM, zero-shot Gemma, zero-shot BERT, and zero-shot GPT-4. For SPEAR, we used BERT Base-110M and Gemma-2b as base models, concatenating soft prompts with token embeddings and training for 10 epochs using cross-entropy loss, AdamW optimizer, and a linear learning rate scheduler. The LSTM baseline employed a two-layer architecture with 128 hidden units, trained for 10 epochs (batch size 32) using Adam (learning rate 0.001) and Cross Entropy Loss, with inputs including day and value features. Zero-shot Gemma and BERT were used without task-specific training to establish baseline performance. Additionally, we evaluated GPT-4 in a zero-shot setting to assess state-of-the-art LLM performance on the task. Performance for all models was evaluated on validation data after each epoch, with final accuracy reported on the test set.

\subsection{Hyperparameters Selection}
\label{app:hyperparam}

\begin{figure*}[tp]
\centering
\begin{subfigure}[b]{.65\columnwidth}
  \centering
  \includegraphics[width=1.0\linewidth, height=3.5cm]{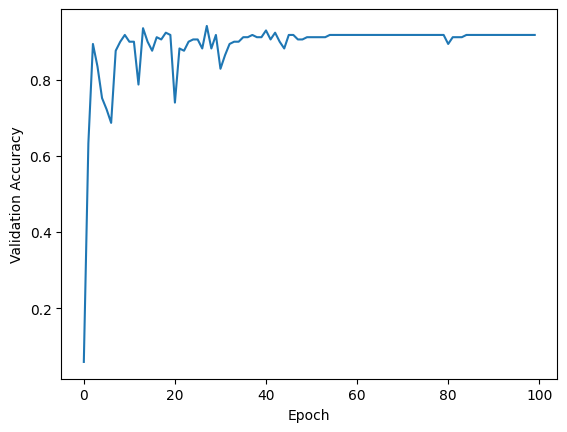}
  \caption{Validation Accuracies}
\end{subfigure}%
\vspace{0.2cm}
\hfill
\begin{subfigure}[b]{.65\columnwidth}
  \centering
  \includegraphics[width=1.0\linewidth, height=3.5cm]{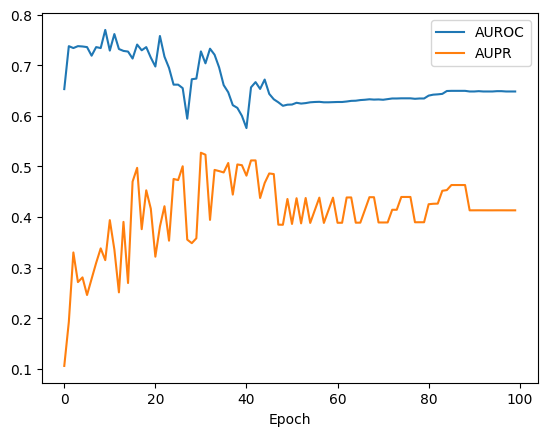}
  \caption{AUROC and AUPR}
\end{subfigure}
\vspace{0.2cm}
\hfill
\begin{subfigure}[b]{.65\columnwidth}
  \centering
  \includegraphics[width=1.0\linewidth, height=3.5cm]{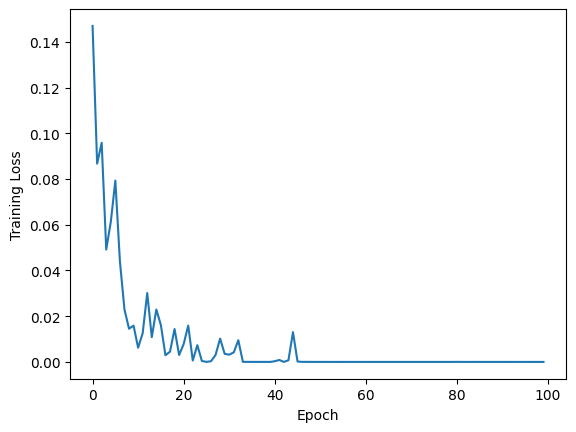}
  \caption{Training Loss}
\end{subfigure}%
\caption{Validation Accuracy, AUROC, AUPR, and Training Loss over 100 Epochs for NASA Dataset}
\label{fig:ch5_AD_NASA_hyperparam}
\end{figure*}

\begin{figure*}[tp]
\centering
\begin{subfigure}[b]{0.65\columnwidth}
  \centering
  \includegraphics[width=\linewidth, height=3.5cm]{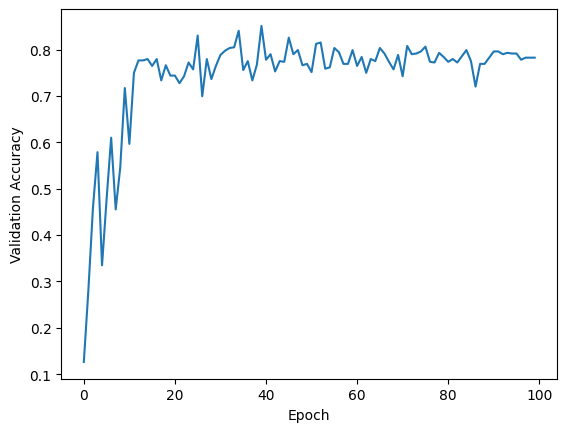}
  \caption{Validation Accuracies}
\end{subfigure}
\vspace{0.2cm}
\hfill
\begin{subfigure}[b]{0.65\columnwidth}
  \centering
  \includegraphics[width=\linewidth, height=3.5cm]{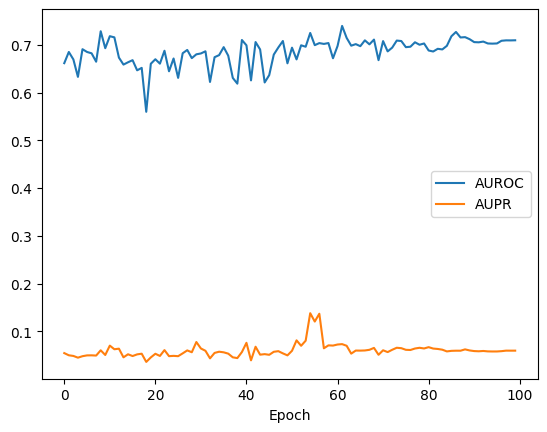}
  \caption{AUROC and AUPR}
\end{subfigure}
\vspace{0.2cm}
\hfill
\begin{subfigure}[b]{0.65\columnwidth}
  \centering
  \includegraphics[width=\linewidth, height=3.5cm]{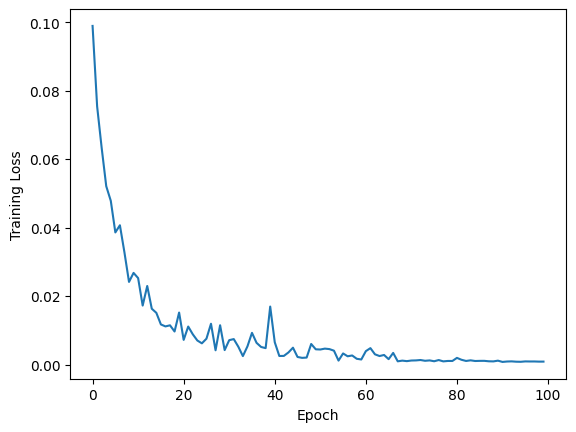}
  \caption{Training Loss}
\end{subfigure}
\caption{Validation Accuracy, AUROC, AUPR, and Training Loss over 100 Epochs for NAB AWS Dataset}
\label{fig:ch5_AD_NAB_hyperparam}
\end{figure*}

Two key hyperparameters were tuned for SPEAR: training epochs and soft prompts embedding size.
Figure~\ref{fig:ch5_AD_NASA_hyperparam} and Figure~\ref{fig:ch5_AD_NAB_hyperparam} show validation metrics over 100 epochs. Accuracy increases significantly in the first 10-20 epochs, with marginal improvements thereafter. AUROC and AUPR stabilize around 40 epochs, with AUPR improvements being particularly important given the highly imbalanced test datasets. Training loss decreases rapidly in early epochs, reaching low values after 30-40 epochs. Based on these convergence patterns, 40 epochs was selected for all experiments.

Table~\ref{tab:virtual_token_table} evaluates soft prompt sizes (10, 20, 30) on MIMIC-IV and NASA datasets. The learnable soft prompt embeddings serve as a guide for the transformer's attention mechanism for time series structure, particularly valuable for contextual anomaly detection. Results show that increasing token size from 10 to 20 notably improves accuracy by capturing more complex temporal dependencies. However, further increasing to 30 yields marginal gains or slight performance degradation, likely due to overparameterization. Therefore, a soft prompt size of 20 was selected as the optimal balance between expressiveness and generalization.

\subsection{Experiment settings}

\subsubsection{\textbf{SPEAR}} 
We used BERT and Gemma 2b as our base model for soft prompts for the time series anomaly detection task. The soft prompts are concatenated with the token embeddings. We then train it for 40 epochs using a cross-entropy loss, a soft prompt size of 20, an AdamW optimizer, and a linear learning rate scheduler. Performance is evaluated on validation data after each epoch, with the final accuracy reported on the test set.
    
\subsubsection{\textbf{LSTM}} We implemented a two-layer LSTM with 128 hidden units to classify time series data. Inputs include two features (day and value). The final LSTM output is combined with the test embeddings for binary classification. The model was trained for 10 epochs (batch size 32) using Adam (learning rate of 0.001) and Cross Entropy Loss, with sequence lengths handled via padding.
    
\subsubsection{\textbf{Gemma, BERT zero-shot}} As baselines, we used Gemma and BERT for zero-shot anomaly detectors with no task-specific training at all. These methods allow us to evaluate and compare the performance against our fine-tuned methodology.
    
\subsubsection{\textbf{GPT4 zero-shot}} We used GPT4 in a zero-shot setting for time series anomaly detection. This allows us to evaluate how a state-of-the-art LLM performs on the same task.

\subsection{Performance} 

\subsubsection{SPEAR vs. Zero-Shot Methods} SPEAR methodology, particularly SPEAR-BERT, consistently outperforms zero-shot approaches across all datasets. This is expected since zero-shot methods rely solely on pre-trained knowledge without task-specific adaptation. Given the extreme imbalance in NASA and NAB datasets, accuracy alone can be misleading. Zero-shot methods achieve high recall but suffer from low precision and F1-scores, producing many false positives. This is problematic for datasets like NASA where anomalies indicate critical spacecraft issues. SPEAR-BERT excels in both AUROC and AUPR, crucial for imbalanced datasets as it accurately identifies minority classes. The significant improvement demonstrates that soft prompts effectively guide LLMs to better detect time series anomalies. It is also worth noting that while zero-shot models such as GPT-4, Gemma, and BERT achieve almost very high on NASA and NAB datasets, their precision remains extremely low. This suggests that these models tend to overpredict anomalies, classifying most, if not all, points as anomalies. This behaviour can be attributed to the lack of task-specific guidance, such as soft prompts or fine-tuning; these models lack the calibration needed to distinguish between the normal and the abnormal. In applications like spacecraft telemetry, this kind of overprediction can lead to excessive false positives, undermining the system's reliability. These findings highlight the importance of task adaptation methods like soft prompting to achieve a better trade-off between recall and precision in anomaly detection.

\subsubsection{SPEAR-BERT vs. SPEAR-Gemma}
SPEAR-BERT consistently outperforms SPEAR-Gemma across all datasets and metrics. On the NASA dataset, SPEAR-BERT achieves 0.9112 validation accuracy compared to SPEAR-Gemma's 0.7515, with more balanced AUROC and AUPR scores across all datasets. BERT's masked language modeling pre-training may make it more receptive to soft prompt modifications, while larger models like Gemma may require more extensive fine-tuning. Additionally, larger models are more prone to overfitting with limited trainable parameters, suggesting that soft prompts alone may be insufficient for optimal performance in larger architectures.

\subsubsection{SPEAR vs. LSTM} On MIMIC-IV, SPEAR-BERT outperforms LSTM, likely due to the dataset's variable-length sequences from irregular medical records. While LSTM requires fixed-size inputs and padding (introducing noise), SPEAR leverages transformer flexibility to handle variable lengths effectively, preserving temporal dynamics essential for detecting context-based anomalies. For NASA dataset, LSTM performs well on accuracy and AUROC but struggles with minority class detection (lower recall and AUPR). SPEAR-BERT offers balanced performance with higher recall and AUPR, making it superior for imbalanced datasets. On NAB dataset, LSTM shows high accuracy but poor F1-score, recall, and precision, indicating majority class bias. SPEAR-BERT achieves more balanced performance with higher F1-score and AUROC. Overall, SPEAR-BERT consistently provides more balanced results across datasets, particularly for imbalanced scenarios common in medical and anomaly detection applications.

\subsubsection{Overall Performance on Different Datasets}
SPEAR consistently outperforms zero-shot base LLMs across all datasets. However, in highly imbalanced datasets like NASA and NAB AWS, accuracy alone is insufficient—AUROC and AUPR must be considered for comprehensive evaluation. The NAB AWS dataset shows low AUPR across all methods despite good accuracy and AUROC, indicating poor minority class performance. This suggests that T-SMOTE may not be optimal for balancing this particular dataset, despite its general improvements over SMOTE. Preprocessing steps—including scaling, quantization, and class imbalance handling—were crucial for adapting continuous data to discrete token-based LLM inputs. These steps, combined with SPEAR methodology, demonstrate promising results for anomaly detection across diverse domains.\newline


\begin{table}[ht]
  \caption{Effect of Token Size on Accuracy on the SPEAR-BERT Performance}
  \label{tab:virtual_token_table}
  \centering
  \begin{tabular}{cccc}
    \toprule
    Soft Prompt Size  & MIMIC IV & NASA  \\
    \midrule
    10             & 0.8900  & 0.7814  \\
    20              & 0.9300 & 0.9112 \\
    30              & 0.9150  & 0.9016 \\
    \bottomrule
  \end{tabular}
\end{table}

\subsubsection{Memory Usage and Parameters Comparison} 
In Fig.~\ref{fig:soft_prompt_sizes}A, we compare the sizes of BERT soft prompt (0.06 MB), Gemma soft prompt (0.16 MB), and LSTM (0.29 MB) models. Soft prompts demonstrate its computational efficiency over traditional approaches like LSTM. By only adding a small number of trainable parameters to pre-trained model, SPEAR enables task-specific adaptation with minimal memory overhead. The BERT and Gemma soft prompts are substantially smaller than the LSTM model. The compact design of the soft prompts allows them to be adaptable to LLMs of different sizes, meaning the same approach could be applied on different LLMs, easing the burden on the users to come up with prompts themselves. As base models increase in size (BERT to Gemma), soft prompts maintain their efficiency advantage, scaling well with larger large language models. In Fig.~\ref{fig:soft_prompt_sizes}B, the total parameters between SPEAR(BERT-110M), SPEAR(Gemma-2b), and GPT-4 are compared. Our approach adds minimal overhead to existing pre-trained LLMs and allows them to outperform frontier LLMs such as GPT-4.

\section{Conclusion}

In this work, we introduced SPEAR, an approach for time series anomaly detection leveraging soft prompts to guide LLMs. Our method demonstrates advantages over zero-shot LLM performances. Particularly in the MIMIC-IV dataset, where the input sequences are irregular, outperforming traditional models like LSTM in those scenarios due to LLMs' flexibility with variable input lengths. Our results highlight the potential of LLMs when used in time series anomaly detection, especially when combined with soft prompts, which helps smaller models such as BERT to outperform larger models such as Gemma, GPT and traditional baselines. 

In our experiments, we focused on smaller foundation models—BERT-Base and Gemma-2B—due to their compatibility with our available computing resources, one NVIDIA A100 GPU. This setup allowed us to train soft prompts without the need for extensive hardware. We acknowledge, however, that the choice of base model can greatly impact anomaly detection performance. As part of future work, we plan to evaluate larger models such as Llama 3/4 and DeepSeek, as well as fine-tuned domain-specific LLMs, which may offer stronger contextual understanding and improved accuracy.

Overall, our study offers insights into how soft prompts and LLMs can be adapted for complex and real-world time series tasks. This paves the way for wider adoption of LLM-based anomaly detection solutions that are both efficient and performance-driven.

\bibliographystyle{IEEEtran}
\bibliography{references}

\end{document}